\documentclass[conference]{IEEEtran}
\IEEEoverridecommandlockouts
\usepackage{cite}
\usepackage{amsmath,amssymb,amsfonts}
\usepackage{algorithmic}
\usepackage{graphicx}
\usepackage{textcomp}
\usepackage{xcolor}
\def\BibTeX{{\rm B\kern-.05em{\sc i\kern-.025em b}\kern-.08em
    T\kern-.1667em\lower.7ex\hbox{E}\kern-.125emX}}
\usepackage{cite}
\usepackage{commath}
\usepackage{subcaption}
\usepackage[noabbrev,capitalise]{cleveref}
\usepackage{acro}

\DeclareAcronym{ml}{
  short = ML,
  long  = Machine Learning
}
\DeclareAcronym{dt}{
  short = DT,
  long  = Digital Twin
}
\DeclareAcronym{fem}{
  short = FEM,
  long  = Finite Element Modeling
}
\DeclareAcronym{lstm}{
  short = LSTM,
  long  = Long Short-Term Memory
}
\DeclareAcronym{doe}{
  short = DoE,
  long  = Design of Experiment
}
\DeclareAcronym{fea}{
  short = FEA,
  long  = Finite Element Analysis
}
\DeclareAcronym{mse}{
  short = MSE,
  long  = Mean Square Error
}
\DeclareAcronym{mae}{
  short = MAE,
  long  = Mean Absolute Error
}
\begin{document}

\title{Identifying Simulation Model Through Alternative Techniques for a Medical Device Assembly Process}

\author{\IEEEauthorblockN{Fatemeh Kakavandi}
\IEEEauthorblockA{\textit{Department of Electrical and Computer Engineering} \\
\textit{Aarhus University}\\
Aarhus, Denmark \\
fateme.kakavandi@ece.au.dk}
}

\maketitle

\begin{abstract}
This scientific paper explores two distinct approaches for identifying and approximating the simulation model, particularly in the context of the snap process crucial to medical device assembly. Simulation models play a pivotal role in providing engineers with insights into industrial processes, enabling experimentation and troubleshooting before physical assembly. However, their complexity often results in time-consuming computations.

To mitigate this complexity, we present two distinct methods for identifying simulation models: one utilizing Spline functions and the other harnessing Machine Learning (ML) models. Our goal is to create adaptable models that accurately represent the snap process and can accommodate diverse scenarios. Such models hold promise for enhancing process understanding and aiding in decision-making, especially when data availability is limited.
\end{abstract}

\begin{IEEEkeywords}
Machine Learning, Simulation, Identification, Snap process
\end{IEEEkeywords}

\section{Introduction}
Simulation models play a crucial role in providing insights to engineers working in industrial processes. They enable various experiments and troubleshooting before building the assembly machine. However, running these simulations can be time-consuming due to the complexity of the computational procedures involved.

To tackle this complexity, we propose an approach that involves conducting key simulation experiments and developing a system identifier to represent the simulation model. In our study, we explore two distinct techniques for identifying the simulation model. The first approach employs spline functions, which are mathematical polynomials. The second approach leverages \ac{ml} models for identification.

By using these techniques, we aim to create a versatile and computationally efficient model that accurately represents the dial snap process and adapts to different scenarios. This approach enhances our understanding of the process and supports informed decision-making when dealing with new input configurations. Moreover, these models can generate synthetic samples and augment training datasets, especially when data availability is limited, as is often the case during process development.

The structure of this paper is organized as follows: We begin by presenting an overview of related work in the field in \cref{sec:sim_related}. Subsequently, we delve into the methodology, which includes introducing the simulation model and its components, as well as the methods for approximating and identifying the simulation model in \cref{sec:sim_meth}. An evaluation of the methodology's performance is then presented in \cref{sec:sim_res}. Finally, we discuss the results and offer insights into future applications \cref{sec:sim_dis}. 

\section{Related work}
\label{sec:sim_related}
Simulation models, as emphasized by \cite{wynn2018process}, are potent tools with versatile applications in the field of manufacturing, especially during the design phase. These models, which draw from a variety of modeling and simulation methods, offer multiple advantages, as outlined by \cite{sinha2001modeling}. They simplify intricate systems while preserving vital variable relationships, thus enabling a deeper comprehension of the system's dynamics. Additionally, they facilitate cost-effective experimentation, particularly in scenarios where real-world testing is either impractical or prohibitively expensive, especially in the early design stages. Simulation models play a pivotal role in optimizing systems, aiding in the identification and comparison of alternatives to enhance overall performance \cite{simbenefit,simbenefitmanufactur}. Moreover, they are invaluable during the design phase, permitting cost-effective adjustments based on simulation outcomes, and are equally crucial for enhancing existing facilities through equipment upgrades or other enhancements. Simulation expedites process improvement by enabling swift "what if" experiments and cost-effective trials on \ac{dt}, iteratively refining the model as real-world data becomes available \cite{compsimbetter,simusecases}.

Nonetheless, it is imperative to acknowledge that the execution of these simulation models can be time-consuming, primarily due to the inherent complexity of the computational procedures involved. Consequently, alternative methodologies have been proposed to approximate these simulation models. One such approach involves fitting a curve to the data generated through simulation, commonly referred to as curve fitting. This process yields an interpretable function that highlights the impact of each input parameter. In this context, splines, a type of mathematical function widely used in numerical analysis for data interpolation and approximation \cite{schumaker2007spline}, are particularly valuable. The primary objective of splines is to determine a smooth curve that best fits a given set of data points. For a practical illustration, refer to \cite{hua2023global}, where splines were employed to approximate linear toolpaths, resulting in improved machining quality and efficiency in CNC machining. Furthermore, the utility of spline functions for approximating simulation models and other datasets is explored in \cite{xia2015second, navarro2023constrained}.

Another approach entails the application of \ac{ml} models to approximate the data generated by simulation models \cite{koziel2020basics, legaard2023constructing}. For instance, in \cite{lazzara2022surrogate}, a model consisting of an \ac{lstm} autoencoder-based architecture appended with a fully connected neural network is employed to predict aircraft dynamic landing responses over time, conditioned by an exogenous set of design parameters. \ac{fem} represents a prevalent technique in simulation modeling, and \cite{kudela2022recent} provides an exhaustive overview of methods and technologies used to approximate \ac{fem}.

Our contribution to the existing body of work lies in introducing a methodology for approximating \ac{fem} within the context of the medical device assembly process. This methodology encompasses two distinct approaches. The first approach leverages curve fitting through spline functions, offering interpretability and the capacity to underscore the significance of various input factors. Conversely, the second approach harnesses deep learning, specifically \ac{lstm} networks, to discern the underlying simulation model. Although the latter approach is considered a black-box model in comparison to the former, it exhibits the ability to discern intricate patterns within the data. This comparative analysis serves as a valuable guide for selecting the most appropriate technique to approximate the simulation model, facilitating informed decision-making in the realm of medical device assembly process modeling and analysis.

\section{Methodology}
\label{sec:sim_meth}
\begin{figure*}[t]
    \centering
    \includegraphics[width=0.8\linewidth]{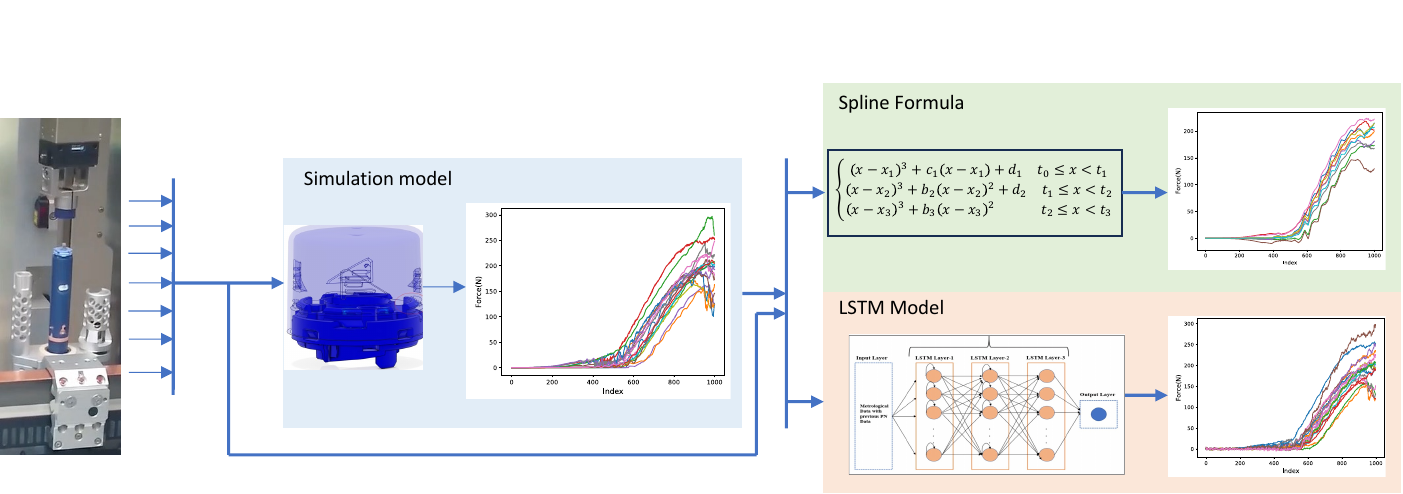}
    \caption{The overview of the methodology employed for approximating the simulation model.}
    \label{fig:sim_overall_meth}
\end{figure*}

This section provides an overview of the methodological steps taken to identify the simulation model, as illustrated in \cref{fig:sim_overall_meth}. We begin by introducing the simulation model, elucidating its input and output components. Following this, we explore two distinct methodologies for approximating and identifying the simulation model.
The initial approach involves the approximation of the simulation model through the definition of a piecewise formula, comprising P spline functions. Conversely, the second approach utilizes a deep learning model to discern underlying patterns within the data generated during simulation, effectively performing an identification process.

\subsection{Dial snap simulation model}
\label{sec:sim_model}
Here, we focus on a simulation model which was developed by our industrial partner using \ac{fea}. The study assesses the impact of various parameters through the utilization of \ac{doe}. We have access to the output data derived from a \ac{doe} investigation conducted for the snap process. This investigation encompasses a range of input parameters, encompassing geometric attributes and assembly variations, while the corresponding outputs consist of force application profiles. The specific input variables, various experiments (Runs), and their corresponding values are comprehensively outlined in \cref{tab:Sim_DoEtable}.

The first two variables, referred to as "Tilt" and "X Offset," pertain to assembly variations, while the remaining variables primarily address geometric modifications. In the initial Run, denoted as "V0," these variables are configured to their nominal values, representing the baseline configuration. Subsequently, in the following Runs, deviations from these nominal values are systematically introduced as part of the \ac{doe} process.

\begin{table*}[t]
	\centering
	\caption{Design of Experiments, variation of parameters.}
	\label{tab:Sim_DoEtable}
	\begin{tabular}{lccccccc}
		\hline
		Run &  Tilt &  X Offset & D out & Wall Thickness & Snap Length & Snap Angle & Snap Width Cut  \\
		\hline \hline 
		V0 & 0 & 0 & 21 & 0.8 & 8.8 & 46.926 & 0 \\ \hline
		V1 & 0 & -0.1 & 20.84 & 0.75 & 8.85 & 48.926 & 0.2 \\ \hline
		V2 & 0 & -0.1 & 21.16 & 0.85 & 8.75 & 44.926 & 0.2 \\ \hline
		V3 & 0 & 0.1 & 20.84 & 0.85 & 8.75 & 48.926 & 0 \\ \hline
		V4 & 0 & 0.1 & 21.16 & 0.75 & 8.85 & 44.926 & 0 \\ \hline
		V5 & 1 & -0.1 & 20.84 & 0.85 & 8.85 & 44.926 & 0 \\ \hline
		V6 & 1 & -0.1 & 21.16 & 0.75 & 8.75 & 48.926 & 0 \\ \hline
		V7 & 1 & 0.1 & 20.84 & 0.75 & 8.75 & 44.926 & 0.2 \\ \hline
		V8 & 1 & 0.1 & 21.16 & 0.85 & 8.85 & 48.926 & 0.2 \\ \hline
		V9 & 1 & 0.1 & 21.16 & 0.85 & 8.75 & 44.926 & 0 \\ \hline
		V10 & 1 & 0.1 & 20.84 & 0.75 & 8.85 & 48.926 & 0 \\ \hline
		V11 & 1 & -0.1 & 21.16 & 0.75 & 8.85 & 44.926 & 0.2 \\ \hline
		V12 & 1 & -0.1 & 20.84 & 0.85 & 8.75 & 48.926 & 0.2 \\ \hline
		V13 & 0 & 0.1 & 21.16 & 0.75 & 8.75 & 48.926 & 0.2 \\ \hline
		V14 & 0 & 0.1 & 20.84 & 0.85 & 8.85 & 44.926 & 0.2 \\ \hline
		V15 & 0 & -0.1 & 21.16 & 0.85 & 8.85 & 48.926 & 0 \\ \hline
		V16 & 0 & -0.1 & 20.84 & 0.75 & 8.75 & 44.926 & 0 \\ \hline
		\hline
	\end{tabular}
\end{table*}

\subsection{Extracted identifier from the simulation model}
\label{sec:sim_identifier}
Creating new simulation models for different sets of input parameters can be computationally burdensome, prompting the need for a more efficient and adaptable modeling approach. To tackle this challenge, we employ two distinct techniques for identifying the simulation model, aiming to uncover the function that maps input parameters to the corresponding outputs. 

In the first approach, we devise a P spline formula, which furnishes us with a concise and precise comprehension of the process. Conversely, the second method involves the application of a \ac{ml} model, capable of learning from and interpreting the simulation data. This approach harnesses the capabilities of advanced algorithms to scrutinize patterns within the data and make predictions regarding the behavior of the dial snap process under varying input parameters. Further elaboration on these two approaches is provided in the subsequent subsections.

\subsubsection{Derived Formula for Dial snap process}
\label{sec:sim_spline}
To obtain an accurate representation of the simulation model, we employ JMP, a predictive analytics software currently in use by our industrial partner. JMP offers a versatile array of tools and techniques that greatly facilitate our analysis. We utilize the P Spline function to fit the input parameters to the force profile over time.

The P Spline, short for Penalized Spline, proves particularly advantageous when dealing with noisy data. It is a specialized variant of the spline function, defined piecewise using polynomials, and capable of smoothly connecting a curve through n+1 points, such as $(x_0,y_0), \ldots, (x_n,y_n)$.

\begin{equation}
	p_i(x) = a_i(x - x_i)^3 + b_i(x - x_i)^2 + c_i(x - x_i) + d_i
	\label{eq:Sim_spline}
\end{equation}

The spline in \cref{eq:Sim_spline} represents a cubic polynomial spline function within the $i$-th interval. The spline function is piecewise-defined, with each interval having its unique set of coefficients: $a_i$, $b_i$, $c_i$, and $d_i$. Here, $x$ represents the input variable or the independent variable at which the spline function is evaluated, while $x_i$ denotes the starting point or knot of the $i$-th interval. The coefficients $a_i$, $b_i$, $c_i$, and $d_i$ are constants specific to each interval and dictate the shape of the cubic polynomial curve within that interval. The terms $(x - x_i)^3$, $(x - x_i)^2$, $(x - x_i)$, and $1$ serve as the basis functions for the cubic polynomial spline, allowing it to capture complex curves and smoothly connect data points within each interval. Notably, the cubic spline function $p_i(x)$ remains valid exclusively within the $i$-th interval, defined by $[x_i, x_{i+1})$. In other words, the spline exhibits piecewise continuity, and its behavior can vary between intervals.

The P Spline formula can efficiently approximate the simulation model and provide insight into understanding the relation of various input to the force profile output. This formula also is interpretable and makes it possible to employ it in industry.

\subsubsection{Machine learning identifier}
\label{sec:sim_ml_iden}
To leverage the capabilities of robust \ac{ml} models for the identification of dynamic systems, we employ an \ac{lstm} network. \ac{lstm}s are highly effective models capable of predicting sequence patterns. In this context, the \ac{lstm} network takes the \ac{doe} variables as input and generates a comprehensive force profile corresponding to these parameters as output.

The final design of the model incorporates multiple \ac{lstm} layers to capture intricate temporal dependencies within the dataset. The use of stacked layers empowers the model to learn hierarchical features. Each layer comprises 250 hidden units, ensuring the model's ability to faithfully replicate the complex patterns present in the simulation data. To predict the force profile sequence, a fully connected layer with 500 output units is utilized.

For the training of the \ac{lstm} model, we partition the dataset of 17 samples into training and testing sets using an 80\% to 20\% split ratio. Each training sample includes seven input parameters and a corresponding force profile comprising 500 data points. The \ac{mse} loss function quantifies the disparity between the predicted force profiles and the ground truth data.

To prevent overfitting and promote generalization, early stopping is implemented, with a threshold of $1e^{-6}$ applied to the loss function. When the loss approaches this minute value, the training process is halted to prevent further iterations. Additionally, the model is trained using the Adam optimizer with a learning rate set to $0.0001$.

\section{Results and evaluation}
\label{sec:sim_res}
\begin{figure*}[t]
	\centering
	\begin{subfigure}[b]{0.30\textwidth}
		\includegraphics[width=0.9\textwidth]{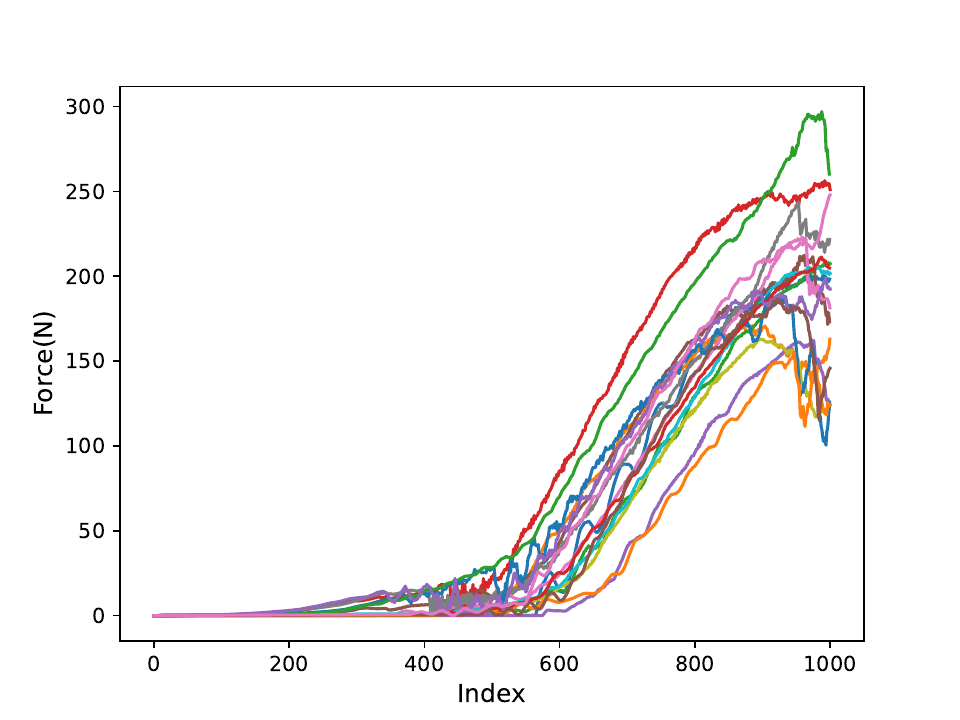}
		\caption{Simulation model.}
		\label{fig:Sim_doe_generated}
	\end{subfigure}
	\begin{subfigure}[b]{0.30\textwidth}
		\includegraphics[width=0.9\textwidth]{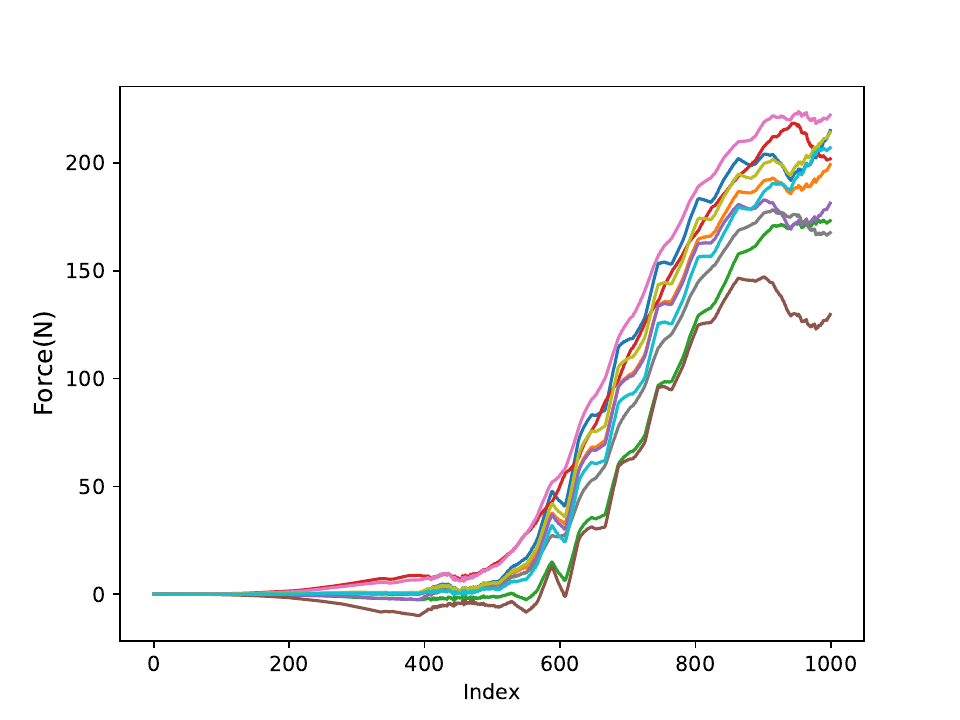}
		\caption{P Spline formula.}
		\label{fig:Sim_spline_generated}
	\end{subfigure}
	\begin{subfigure}[b]{0.30\textwidth}
		\includegraphics[width=0.9\textwidth]{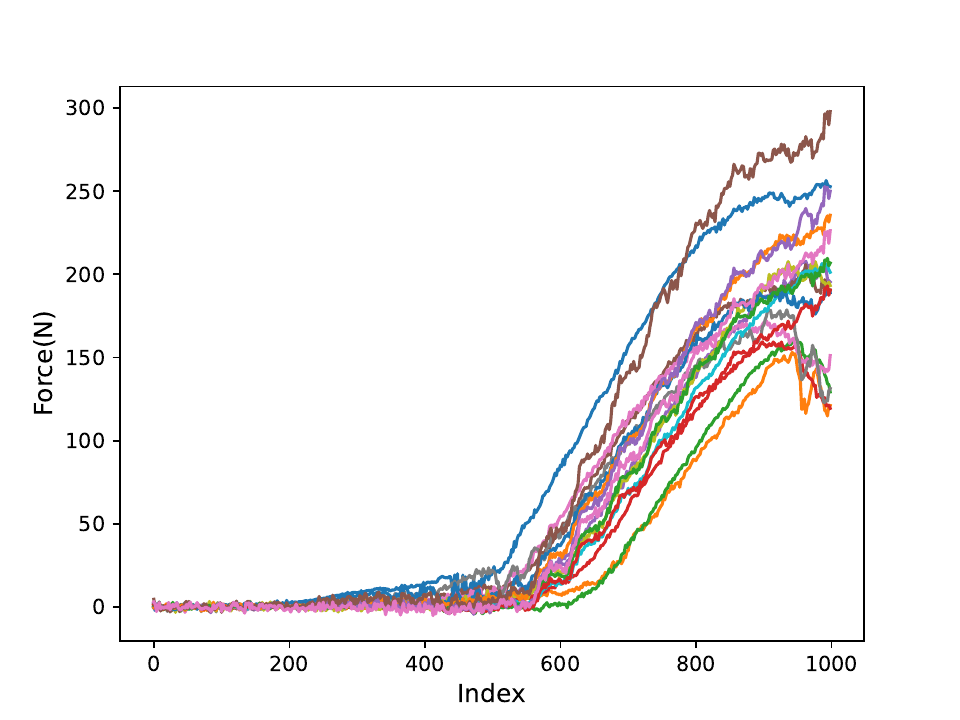}
		\caption{\ac{lstm} model.}
		\label{fig:Sim_ML_generated}
	\end{subfigure}
	\caption{Generated samples via different techniques.(\subref{fig:Sim_doe_generated}) shows the data generated via simulation model through \ac{doe}. (\subref{fig:Sim_spline_generated}) data generated through P Spline function. (\subref{fig:Sim_ML_generated}) illustrates the data generated via \ac{lstm}.}
	\label{fig:sim_all_data}
\end{figure*}

The data resulting from two distinct methods, as elaborated in \cref{sec:sim_identifier}, are presented in \cref{fig:sim_all_data}. As expected, the utilization of the P Spline function yields data characterized by smoothness, effectively mitigating simulation noise. Moreover, the generated patterns closely resemble the data obtained from the simulation model. On the other hand, the \ac{lstm} model demonstrates impressive proficiency in capturing force profile patterns based on input parameters. This is clearly illustrated in \cref{fig:sim_all_data} (\subref{fig:Sim_ML_generated}), where the samples generated by the \ac{lstm} model faithfully replicate the existing patterns in the simulation data.

To evaluate the performance of the Spline formula in representing the simulation data, we compute the \ac{mae}. The results, presented in \cref{tab:Sim_performance}, provide insights into the effectiveness of the two employed techniques for approximating the simulation model. Notably, the \ac{lstm} model outperforms the Spline formula, as evidenced by its lower \ac{mae} value.

These two approaches offer distinct methods for approximating the simulation model. The first method, which employs the Spline formula, provides an interpretable approach. In contrast, the second approach, utilizing the \ac{lstm}, demonstrates superior precision. The availability of these approximators streamlines the execution of the model for new sets of input variables, reducing effort and making it accessible to individuals with diverse backgrounds. These approaches are not only more time-efficient but also enhance the understanding of the underlying process.

\begin{table}[t]
	\centering
	\caption{\ac{mae} for two approaches that approximate the simulation model.}
	\label{tab:Sim_performance}
	\begin{tabular}{lc}
		\hline
		Method & \ac{mae}\\
		\hline \hline
		Spline formula & $3.0176$\\
		\ac{ml} model &$1.357$ \\
		\hline
	\end{tabular}
\end{table} 

\section{Discussion}
\label{sec:sim_dis}
The evaluation of the employed approaches underscores the value of developing alternative models capable of approximating the Dial Snap simulation model. The first approach, employing the Spline formula, offers an interpretable representation of the impact of input variables, while the \ac{lstm} model places a stronger emphasis on accurately reproducing the data generated by the simulation model. The utilization of these two distinct techniques can provide valuable insights for decision-making. For instance, when interpretability is paramount, the Spline formula can be the preferred choice, while for greater precision, the \ac{lstm} model may be more suitable.

However, it's important to acknowledge certain limitations. The \ac{lstm} model relies heavily on a substantial volume of data, which can pose challenges in scenarios with limited data availability. Additionally, both the simulation model and its approximators are reliable only within the validation range, limiting their extrapolation capabilities.

Furthermore, the formula we derived in Section \ref{sec:sim_spline} serves as a valuable tool for generating synthetic force profiles, which holds significant relevance for future predictive applications, as elaborated in the forthcoming chapter. To facilitate this, we intentionally constructed alternative input samples that deviated from the nominal values, while ensuring that these variables remained within an acceptable range around the nominal values.

\section{Conclusion}
\label{sec:sim_conclusion}
In this scientific paper, we have presented two distinct techniques for identifying and approximating a simulation model for the dial snap process in medical device assembly. We have demonstrated the use of P Spline functions to create an interpretable formula that captures the relationship between input parameters and the force profile generated by the simulation model. Additionally, we have applied \ac{lstm} deep learning models to predict the force profile based on input parameters.

Our evaluation indicates that both approaches have their merits. The Spline formula offers interpretability and a clear understanding of the impact of input parameters on the simulation model. On the other hand, the \ac{lstm} model provides higher accuracy in replicating the simulation data. The choice between these approaches may depend on the specific requirements of the application, with the Spline formula being suitable when interpretability is critical and the \ac{lstm} model being preferred for accuracy.

Overall, these techniques for approximating the simulation model offer valuable tools for engineers and researchers working on the dial snap process. They can streamline the process of experimentation and decision-making, ultimately leading to more efficient and informed development of medical device assembly processes.

\bibliography{references}

\end{document}